\documentclass[11pt, a4paper, logo, copyright, nonumbering]{astribot}
\usepackage[authoryear, sort&compress, round]{natbib}
\usepackage{dblfloatfix}
\usepackage{ulem}
\usepackage{caption}
\usepackage{dramatist}
\usepackage{xspace}
\usepackage{pifont} 
\usepackage{multirow}
\usepackage{tcolorbox}
\usepackage{xltabular}
\usepackage{longtable}
\usepackage{hyperref}
\usepackage{amsfonts}
\usepackage{amsmath}
\usepackage{amssymb}
\usepackage{lineno}
\usepackage{multirow}
\usepackage{adjustbox}
\usepackage{siunitx}
\usepackage{mathtools}
\usepackage{multirow}
\usepackage[bottom]{footmisc}
\usepackage{fontawesome5}
\usepackage{subfigure}
\usepackage{setspace}
\usepackage{lipsum} 
\usepackage{multicol} 

\makeatletter
\def\@BTrule[#1]{%
  \ifx\longtable\undefined
    \let\@BTswitch\@BTnormal
  \else\ifx\hline\LT@hline
    \nobreak
    \let\@BTswitch\@BLTrule
  \else
     \let\@BTswitch\@BTnormal
  \fi\fi
  \global\@thisrulewidth=#1\relax
  \ifnum\@thisruleclass=\tw@\vskip\@aboverulesep\else
  \ifnum\@lastruleclass=\z@\vskip\@aboverulesep\else
  \ifnum\@lastruleclass=\@ne\vskip\doublerulesep\fi\fi\fi
  \@BTswitch}
\makeatother

\addto\extrasenglish{
}

 {\begin{list}{}%
         {\setlength{\leftmargin}{#1}}%
         \item[]%
 }
 {\end{list}}
 
\renewcommand{\today}{}
\newcommand{\acronym}[0]{\mbox{Astribot}\xspace}
\newcommand{\algoname}[0]{\mbox{DuoCore-WB}\xspace}
\newcommand{\fullname}[0]{\mbox{Astribot Suite}\xspace}

\title{ Towards Human-level Intelligence via Human-like Whole-Body Manipulation}
\author[*]{
Astribot Team
\\
\small
\texttt{research@astribot.com}
}

\begin{abstract}
Building general-purpose intelligent robots has long been a fundamental goal of robotics. A promising approach is to mirror the evolutionary trajectory of humans: learning through continuous interaction with the environment, with early progress driven by the imitation of human behaviors. Achieving this goal presents three core challenges: (1) designing safe robotic hardware with human-level physical capabilities; (2) developing an intuitive and  scalable whole-body teleoperation interface for data collection; and (3) creating algorithms capable of learning whole-body visuomotor policies from human demonstrations.
To address these challenges in a unified framework, we propose \fullname, a robot learning suite for whole-body manipulation aimed at general daily tasks across diverse environments. We demonstrate the effectiveness of our system on a wide range of activities that require whole-body coordination, extensive reachability, human-level dexterity, and agility. Our results show that \acronym's cohesive integration of embodiment, teleoperation interface, and learning pipeline marks a significant step towards real-world, general-purpose whole-body robotic manipulation, laying the groundwork for the next generation of intelligent robots.
\end{abstract}

\begin{document}

\maketitle
\newpage

\begin{figure}[h]
\setlength{\linewidth}{\textwidth}
\setlength{\hsize}{\textwidth}
    \centering
\includegraphics[width=\textwidth]{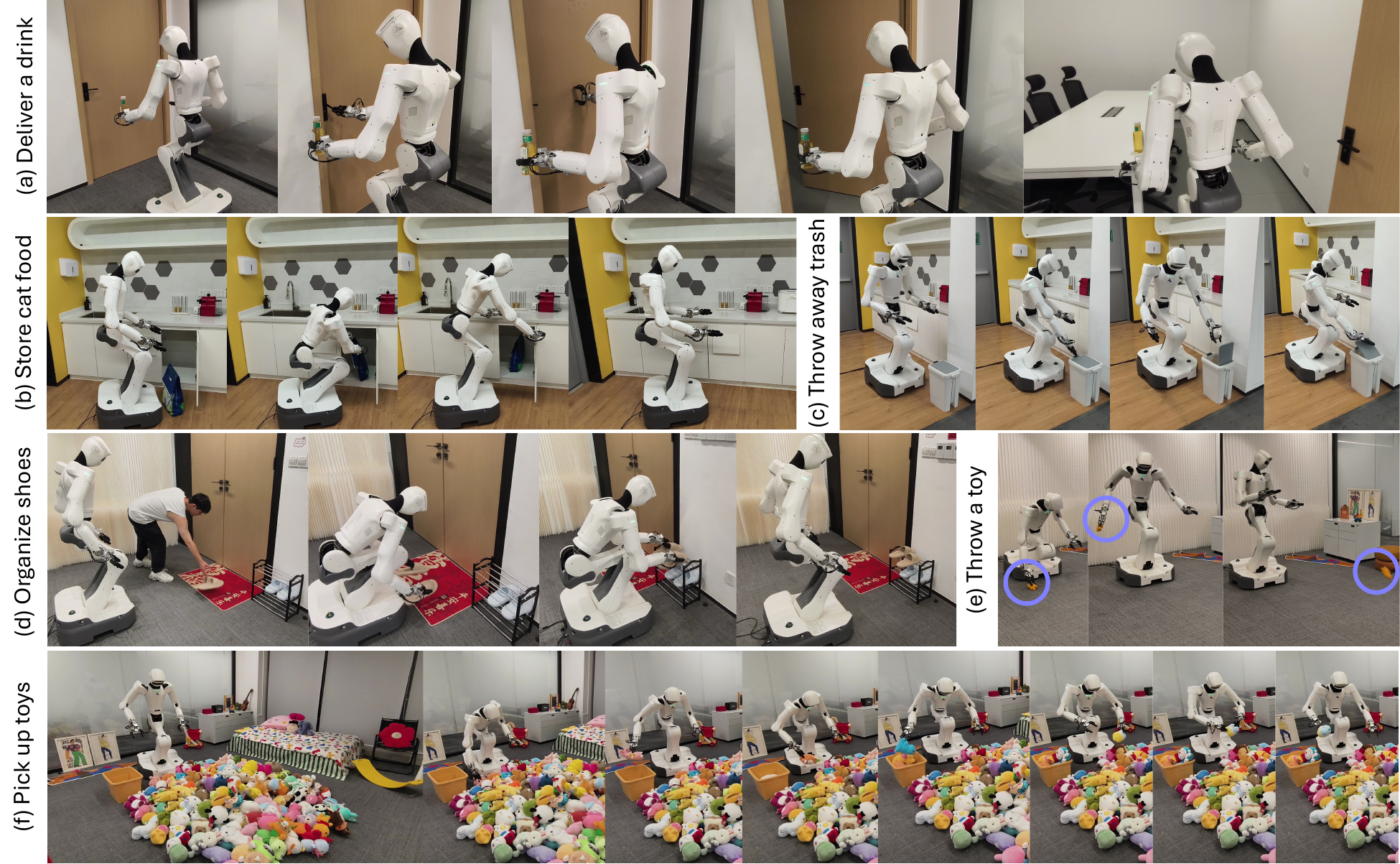}
\caption{\textbf{
\fullname enables a diverse range of everyday activities through whole-body coordination, demonstrating extensive reachability, human-like dexterity, and agility.} Shown are rollout trajectories of \algoname policies, one per task, trained using a simple yet effective imitation learning algorithm on demonstrations collected via our whole-body teleoperation system.
Activities include:
\textbf{(a) Deliver a drink:} The robot grasps a drink with one arm, navigates to the door, opens it with the other arm, enters the room, and places the drink on the desk.
\textbf{(b) Store cat food:} The robot lifts a $\sim$ 2kg bag of cat food using both arms, places it inside a cabinet, and then closes the cabinet door. 
\textbf{(c) Throw away trash:} The robot navigates to the trash bin, opens the lid by pressing a button, discards a used paper cup, and subsequently closes the lid.
\textbf{(d) Organize shoes:} After a person randomly places a pair of shoes on the carpet, the robot approaches the shoes, picks them up with both hands, navigates to the shoe rack, and neatly places them on the rack.
\textbf{(e) Throw a toy:} The robot picks up a toy from the floor and throws it far way.
\textbf{(f) Pick up toys:} The robot collects toys scattered on the ground and places them into a storage bin using both arms. 
}
\label{fig:teaser}
\end{figure}

\section{Introduction}
Developing intelligent robots that can inhabit human environments, perceive the world as we do, understand the consequences of their actions, interact safely with their surroundings, and ultimately co-exist with humans to assist in everyday life has long remained an unrealized vision rooted in science fiction~\citep{gupta2021embodied,team2021creating}. We posit that achieving such human-like capabilities - and ultimately human-level intelligence in robots - requires learning from both diverse and fine-grained interactions with the physical world. A practical and efficient starting point is to imitate human behaviors, with an emphasis on human-like capabilities: whole-body coordination with extensive reachability, dexterity, and agility. 

Achieving autonomous whole-body manipulation in everyday settings entails three fundamental challenges. First, it requires carefully engineered robotic hardware that is not only safe for deployment in human environments but also exhibits human-level physical capabilities and expressiveness. Second, an intuitive and scalable data collection interface is essential, one that enables non-experts to perform a wide variety of daily tasks in a natural and efficient manner. Third, it demands learning models that can capture the intricacies of whole-body control, with the potential to scale and to generalize to open-world environments and activities.

\begin{figure*}[t]
    \centering
    \includegraphics[width=\textwidth]{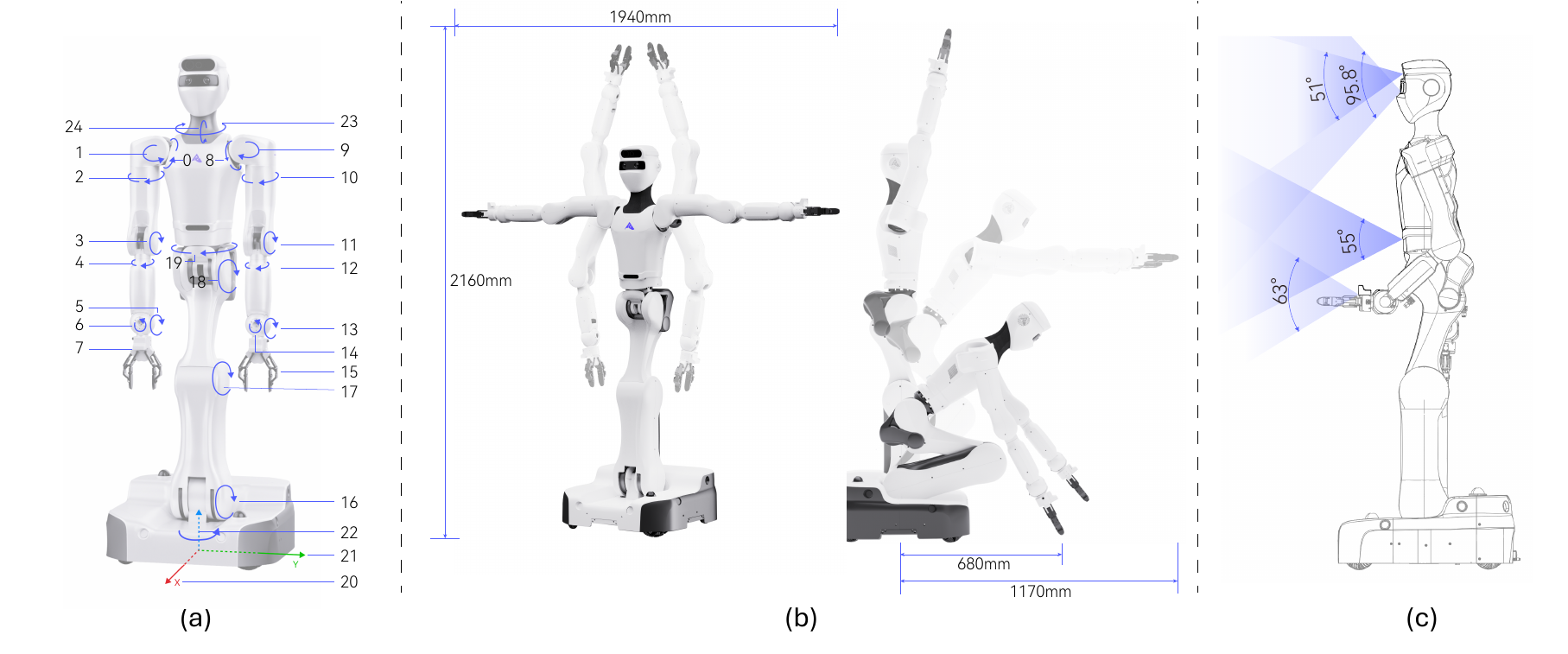}
    \caption{\textbf{Astribot S1 - the robotic platform of \fullname.} Astribot S1 is equipped with two 7-DoF arms, each equipped with a parallel-jaw gripper, as well as a 4-DoF articulated torso, a 2-DoF head, and a 3-DoF omnidirectional mobile base. The system's DoFs are illustrated in (a). Panels (b) and (c) depict the robot's worksapce range of motion, and the distribution of its camera Field of View (FoV), respectively.
    \label{fig:hardware_system}}
\end{figure*}

To this end, we address the aforementioned challenges by introducing a unified framework for learning real-world whole-body manipulation tasks, as exemplified in Fig.~\ref{fig:teaser}. Our framework consists of three core components: First, we develop a highly capable robotic platform - a dual-arm manipulator with a flexible torso and a mobile base - designed to emulate human behaviors effectively and gracefully. Second, we introduce a whole-body teleoperation interface that prioritizes general applicability, high usability, and low latency, enabling scalable data collection by everyday non-expert users. Finally, we present the whole-body visuomotor policy (\algoname) - a simple yet effective learning algorithm that models coordinated whole-body actions using RGB observations and carefully designed action representation.

Our robotic platform is a high-performance, robust, and safe mobile manipulator designed for general tasks. It features an innovative cable-driven design~\citep{qian2018review} that closely emulates human musculature, enabling compliant motion and nuanced force application. Compared to conventional rigid-link robots, our design offers superior payload capacity, reduced backlash and inertia, enhanced compactness, and improved operational safety. Its lightweight structure, low-friction transmission, and soft cushioning characteristics enable high-resolution force control, which is crucial for AI-driven manipulation tasks that require precise force feedback~\citep{dai2025co}. Minimal control latency and high-precision trajectory tracking are achieved via hybrid rigid-soft body dynamics modeling. Furthermore, the transmission mechanisms and manufacturing processes are carefully optimized to minimize surface wear and ensure long-term mechanical durability, satisfying the demands of real-world lifespan requirements.

For whole-body teleoperation, we introduce an intuitive and cost-effective system leveraging an VR headset and handheld joysticks. Hand poses captured by the joysticks are mapped to robot's end-effector positions and orientations, which are subsequently converted into joint commands via whole-body control. The system supports two control modes: (1) a first-person view mode, optimized for precise and complex manipulation tasks as well as remote teleoperation; and (2) a third-person view mode, designed for large-range whole-body motion with minimal transmission latency, making it well-suited for high-dynamic tasks or near-field data collection.

With efficient data collection across diverse tasks, we introduce \algoname, a simple yet effective imitation learning algorithm designed to model whole-body actions. Rather than focusing on architectural complexity, we highlight three core design choices: (1) RGB-based visual perception, enabling the use of pre-trained vision encoders~\citep{dosovitskiy2020image,oquab2023dinov2,radford2021learning,zhai2023sigmoid} for visual generalization, and facilitating scalability through compatibility with large-scale VLA training~\citep{driess2023palm, zitkovich2023rt,kim2024openvla,black2024pi_0,liu2024rdt,team2025gemini, bjorck2025gr00t,intelligence2025pi_}; (2) Whole-body policy control in the end-effector (EE) space using SO(3) as representation of orientation~\citep{geist2024learning}, expressed as deltas in the egocentric frame of each end-effector. Learning in the end-effector space effectively mitigates error accumulation - an issue that becomes particularly pronounced when learning whole-body policies - compared to learning directly in joint space.  Furthermore, representing actions relative to the egocentric frame emphasizes alignment between visual observations and action targets, while being more robust to large viewpoint shifts that frequently occur in complex, coordinated whole-body activities; (3) A real-time trajectory generation module (RTG), which refines predicted action chunks into smooth, accurate, and continuous execution utilizing quadratic programming (QP) optimization. This module acts as a post-processor that is robust, real-time capable, and agnostic to the upstream policy generating the action chunks.

We evaluate \fullname on six representative real-world whole-body tasks as illustrated in Fig.~\ref{fig:teaser}. Each \algoname policy is trained on data collected via our teleoperation interface, with one policy per task. \textbf{Deliver a drink} accesses the ability to execute long-horizon tasks, perform mobile manipulation, and dexterous manipulation with articulated objects (door handle). \textbf{Store cat food} evaluates coordinated bimanual manipulation within constrained spaces (e.g. low cabinets), and dynamic stability while handling high payloads ($\sim$ 2kg). \textbf{Throw away trash} tests bimanual coordination across multi-stage tasks and precise manipulation of articulated objects (a trash bin lid). \textbf{Organize shoes} focuses on whole-body coordination in low-height spaces and the capability for synchronous bimanual manipulation. \textbf{Throw a toy} demonstrates the ability to perform highly dynamic movements and precise grasping of objects from the floor. \textbf{Pick up toys} provides a comprehensive assessment of the ability to continuously manipulate multiple objects accurately with coordinated bimanual interaction. The learned \algoname policies exhibit strong performance, achieving an average success rate of 80\% and a peak success rate of 100\%. We argue that \fullname's integrated robotic embodiment, teleoperation interface, and learning framework represent a significant step towards enabling real-world whole-body manipulation for everyday tasks.

\section{Hardware System}
\label{sec:system}
This section details the hardware components of \fullname. We begin with the design of our mobile dual-arm manipulator as illustrated in Fig.~\ref{fig:hardware_system}, which features two 7-DoF arms, a flexible 4-DoF torso, a 2-DoF head, and a 3 DoF mobile base - optimized for daily tasks that require high mobility and extended end-effector reachability. We then introduce a cost-effective whole-body teleoperation interface that enables intuitive and efficient data collection across a broad range of real-world manipulation scenarios.

\subsection{Robotic Platform}
\begin{table}[]
\centering
\begin{tabular}{|c|c|c|}
\hline
& Astribot S1     & Normal Adult Male                                                              \\ \hline
Single arm DoF                                                                  & 7                                                                    & 7                                                                              \\ \hline
\begin{tabular}[c]{@{}c@{}}Payload per arm (horizontal reach)\end{tabular}   & 5kg                                                                  & 3-5kg                                                                          \\ \hline
\begin{tabular}[c]{@{}c@{}}Arm EE maximum velocity\end{tabular}     & \begin{tabular}[c]{@{}c@{}} $\geq$10 m/s\end{tabular}                    & \begin{tabular}[c]{@{}c@{}} 5-10 m/s (swing arm)\end{tabular}                   \\ \hline
\begin{tabular}[c]{@{}c@{}}Arm EE maximum acceleration\end{tabular} & \begin{tabular}[c]{@{}c@{}}100 m/s²\end{tabular}                  & \begin{tabular}[c]{@{}c@{}}50-100 m/s² (swing arm)\end{tabular}                \\ \hline
\begin{tabular}[c]{@{}c@{}}Arm EE positioning repeatability\end{tabular}            & \begin{tabular}[c]{@{}c@{}}±0.1 mm\end{tabular}                    & \begin{tabular}[c]{@{}c@{}}±1-5 mm (training up to ±0.5 mm)\end{tabular}       \\ \hline
\begin{tabular}[c]{@{}c@{}}Arm EE absolute positioning \\accuracy repeatability\end{tabular}            & \begin{tabular}[c]{@{}c@{}}±1 mm\end{tabular}                    & \begin{tabular}[c]{@{}c@{}}±1-5 mm (training up to ±0.5 mm)\end{tabular}       \\ \hline
Height                                                                          & \begin{tabular}[c]{@{}c@{}}170 cm\end{tabular}                     & 173 cm                                                                         \\ \hline
Total weight (with Battery)                                                                          & \begin{tabular}[c]{@{}c@{}}90 kg\end{tabular}                      & 70 kg                                                                          \\ \hline
Full arm span (with Grippers)                                                                        & \begin{tabular}[c]{@{}c@{}}194 cm\end{tabular}                     & 171-175 cm                                                                     \\ \hline
\end{tabular}
\caption{Comparison of operational parameters between the Astribot S1 robot and an average adult male. Astribot S1 exhibits human-level or superhuman capabilities on on all metrics.}
\label{tab:compare_with_human}
\end{table}
As shown in Fig.~\ref{fig:hardware_system}, the Astribot S1 robot features two 7-DoF arms mounted on a highly articulated 4-DoF torso. Each arm is equipped with a parallel-jaw gripper capable of handling payloads up to \SI{5}{\kilogram}, enabling manipulation of a wide variety of everyday objects. The grippers achieve opening and closing times as fast as 0.15 seconds, enabling data collection on agile, human-like behaviors such as throwing. 
The 4-DoF torso allows for waist rotation, hip flexion, and knee-like articulation, allowing the robot to smoothly transition between standing and squatting postures. This design significantly enhances vertical mobility and extends the effective workspace. The 2-DoF head allows for dynamic gaze control and task-relevant visual focus, mirroring human perceptual behavior in large or cluttered environments.
The S1 robot achieves an effective vertical reach range from ground level to \SI{2}{\meter}, and a horizontal reach of up to \SI{1.94}{\meter} (with grippers).
Designed to operate effectively in human-centric environments and support a diverse range of tasks, Astribot S1 surpasses key human performance metrics, as benchmarked against the average adult male as detailed in table~\ref{tab:compare_with_human}.

For perception, the S1 robot is equipped with a diverse array of onboard sensors to enable robust scene understanding and manipulation. The FoV of the sensors are illustrated in Fig. ~\ref{fig:hardware_system} (c). The head integrates a stereo RGB camera and an RGB-D Orbbec Femto Bolt, offering high-resolution visual input from an elevated viewpoint. Two Intel RealSense D401 RGB-D cameras are mounted on the wrists to provide close-range observations during fine manipulation. These are complemented by an Orbbec Gemini 335 RGB-D camera on the chest for mid-range perception, and a Livox MID-360 LiDAR on the chassis for 360-degree spatial awareness and mapping. All RGB-D cameras operate at \SI{30}{\hertz}.

To support stable and agile navigation, the robot's torso is mounted on an omnidirectional mobile base capable of holonomic motion across the ground plane. The base supports a maximum linear speed of \SI{2}{\meter\per\second}. The chassis integrates four drive modules that enable forward and backward motion, lateral translation, and in-place rotation, providing high maneuverability in constrained environments.

\subsection{Low-latency Teleportation for Scalable Data Collection}
To support seamless control of high DoF mobile manipulators and enable scalable data collection for downstream policy learning, we develop an optimized whole-body teleoperation system designed with the following key considerations: (1) Intuitive and real-time whole-body control to seamlessly coordinate complex movements; (2) Robust safety mechanisms to protect both the operator and the robot; (3) High-quality demonstration capture for effective policy learning; (4) A low-cost implementation to enhance accessibility.

Our teleoperation system leverages the Meta Quest 3S VR headset, mapping the poses of the headset and the handheld joysticks to the robot's end-effectors. A whole-body control (WBC) framework translates these inputs into joint-level commands, enabling real-time motion synchronization between human operator and the robot. The system supports two complementary control modes: (1) \textbf{First-person view mode:} the operator wears the VR headset and controls the robot from its first-person perspective. This setup aligns with natural human perception and operational habits, making the interaction intuitive and lowering the learning barrier. It is particularly suited for remote teleoperation tasks such as service robotics and human-robot collaboration. This control mode supports fine and complex manipulation, such as folding clothes, organizing small items, and precision grasping, providing high control accuracy as well as an immersive experience. (2) \textbf{Third-person view mode:} the operator wears the VR headset on the chest and stands beside the robot, directly observing its full-body state for control. This setup is particularly well-suited for large-range whole-body motion control, especially for dynamic demonstrations and posture adjustments involving high DoFs. Since the operator can directly observe the robot’s behavior in real time without relying on image transmission, this mode effectively eliminates latency-related uncertainties, improving both safety and responsiveness. As a result, it is ideal for executing highly dynamic tasks or performing near-field data collection. The two control modes can be flexibly switched according to task requirements, striking a balance between remote fine manipulation and agile, close-range control. The system is broadly applicable to service robot demonstrations, human-robot collaboration, and large-scale data collection across a wide range of scenarios.

\begin{figure}[t]
  \centering
  \includegraphics[width=\linewidth]{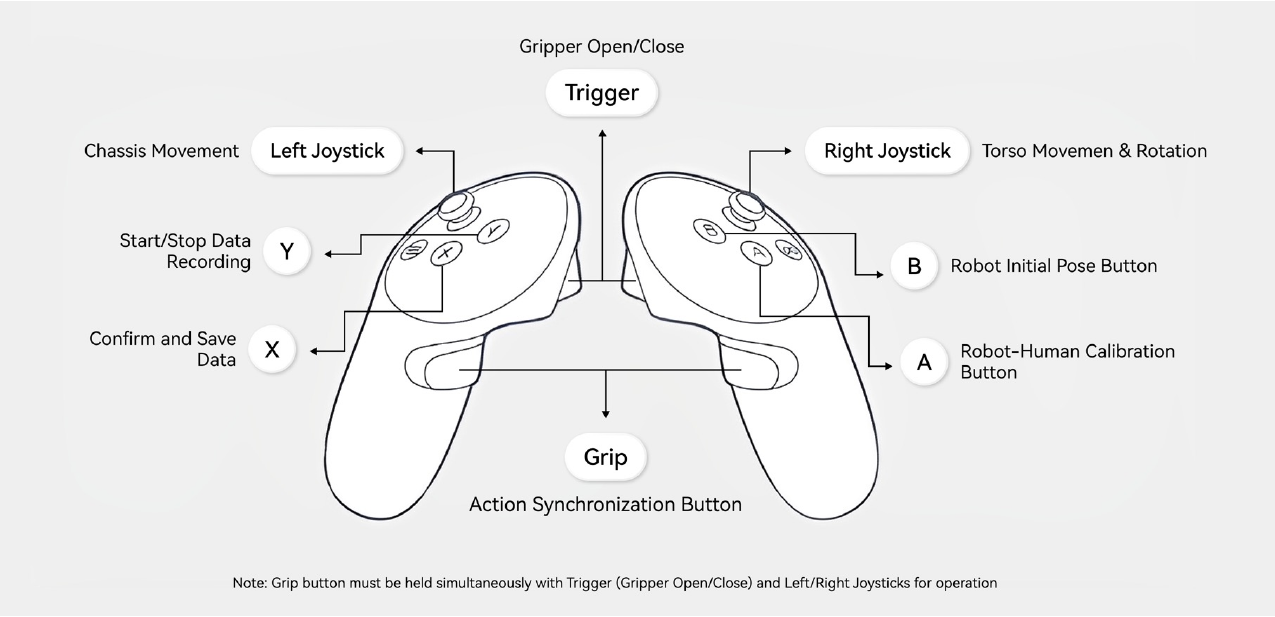}
   \caption{\textbf{Re-targeting joysticks for whole-body teleoperation and demonstration recording.}}
   \label{fig:controller}
\end{figure}

\paragraph{Intuitive and Real-time Whole-Body Control}
As illustrated in Fig.~\ref{fig:controller}, we reconfigure the Meta Quest 3S joysticks to support intuitive teleoperation and demonstration recording. The key commands are as follows:
\begin{itemize}
    \item Holding the grip button activates action-following mode, enabling the robot to track whole-body human movements. 
    \item The triggers control the opening and closing of the grippers.
    \item The left thumbstick issues commands to the mobile base.
    \item The right thumbstick adjusts the vertical positions of the limbs.
\end{itemize}
This configuration enables users to simultaneously control whole-body motion and base navigation in an intuitive and user-friendly manner, supporting efficient and accurate teleoperation. The system is supported by an optimized low-latency control suite. The system operates at a control frequency of 100Hz, and an end-to-end response delay is 20 ms from teleoperation command to the robot action, ensuring smooth and responsive teleoperation. Note that for first-person view mode, image transmission latency is around 100 ms.

\paragraph{Safety}
Ensuring the safety of both the operator and the robot is critical when collecting data for general-purpose tasks, especially in the presence of potential operational errors. The system incorporates two safety features. \textbf{(1) Robot damage protection:} we implement a tipping protection mechanism which constrains the robot's whole-body center of mass, enabling safe operation across diverse configurations and motion speeds, thereby minimizing the risk of falls. Additionally, Cartesian space target tracking is performed while adhering to refined self-collision constraints, including interactions between robot components and grippers. This is achieved through an object proximity distance computation engine integrated with a Whole-Body Control (WBC) controller in Cartesian space~\citep{liu2023collision,jin2024real}; \textbf{(2) Impact force mitigation:} our cable-driven robot design inherently promotes safety due to its mechanical structure with passive compliance. Furthermore, external forces on both the robot body and grippers are estimated using classical model-based algorithms that leverage precise whole-body dynamics and motor feedback signals~\citep{haddadin2017robot}, eliminating the need for additional force sensors. These estimated external forces, combined with high-frequency torque limiting, enable active compliance control across the body and grippers, effectively preventing large impact forces during contact.

\begin{figure*}[t]
    \centering
    \includegraphics[width=\textwidth]{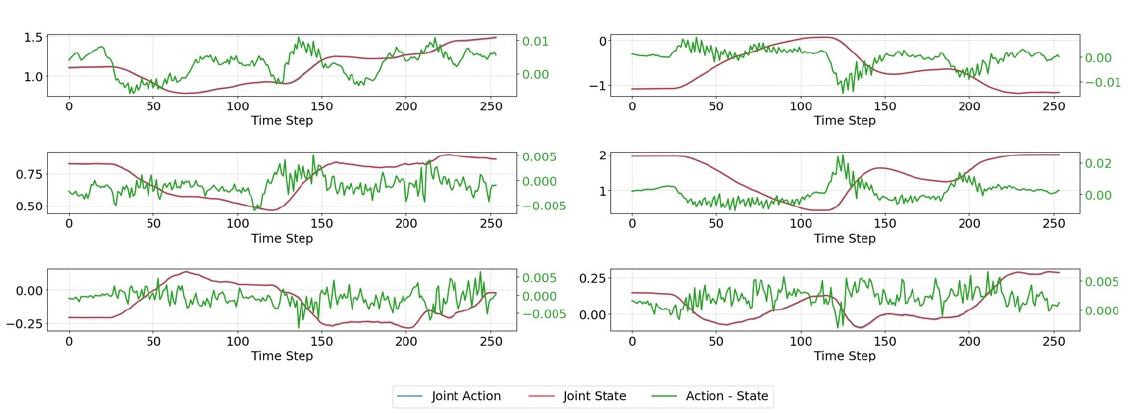}
    \caption{\textbf{High-precision trajectory tracking.} Recorded demonstration for the task ``pick up toys'' via our teleoperation interface. The plot shows data of 6 randomly selected joints. The blue line represents the commanded joint positions, while the red line shows the actual positions. The tracking error is indicated by the dotted green lines, which remain consistently small, demonstrating high-precision trajectory tracking.} 
    \label{fig:tracking_error}
\end{figure*}

\paragraph{High-quality Demonstration}
The choice of teleoperation interface plays a crucial role in the performance of downstream policy learning, particularly in imitation learning, where the ability to consistently replay successful demonstration trajectories is essential. Empirically, our teleoperation system achieves near 100\% replay success rates while maintaining low action tracking errors - both key indicators of demonstration quality and policy reproducibility. As shown in Fig.~\ref{fig:tracking_error}, we visualize tracking errors of six randomly selected joints during demonstration collection of the ``pick up toys'' task (as illustrated in Fig.~\ref{fig:teaser}), where the specific session features picking up a toy with the right arm.

\paragraph{Low Cost}
The teleoperation interface is designed to be highly cost-effective, utilizing only off-the-shelf Meta Quest 3S headset and joysticks. The entire system operates with a total hardware cost of less than 300 USD, making it accessible and scalable for widespread deployment.

\section{Whole Body Policy Learning Method}
\label{sec:method}
\begin{figure*}[t]
    \centering
    \includegraphics[width=\textwidth]{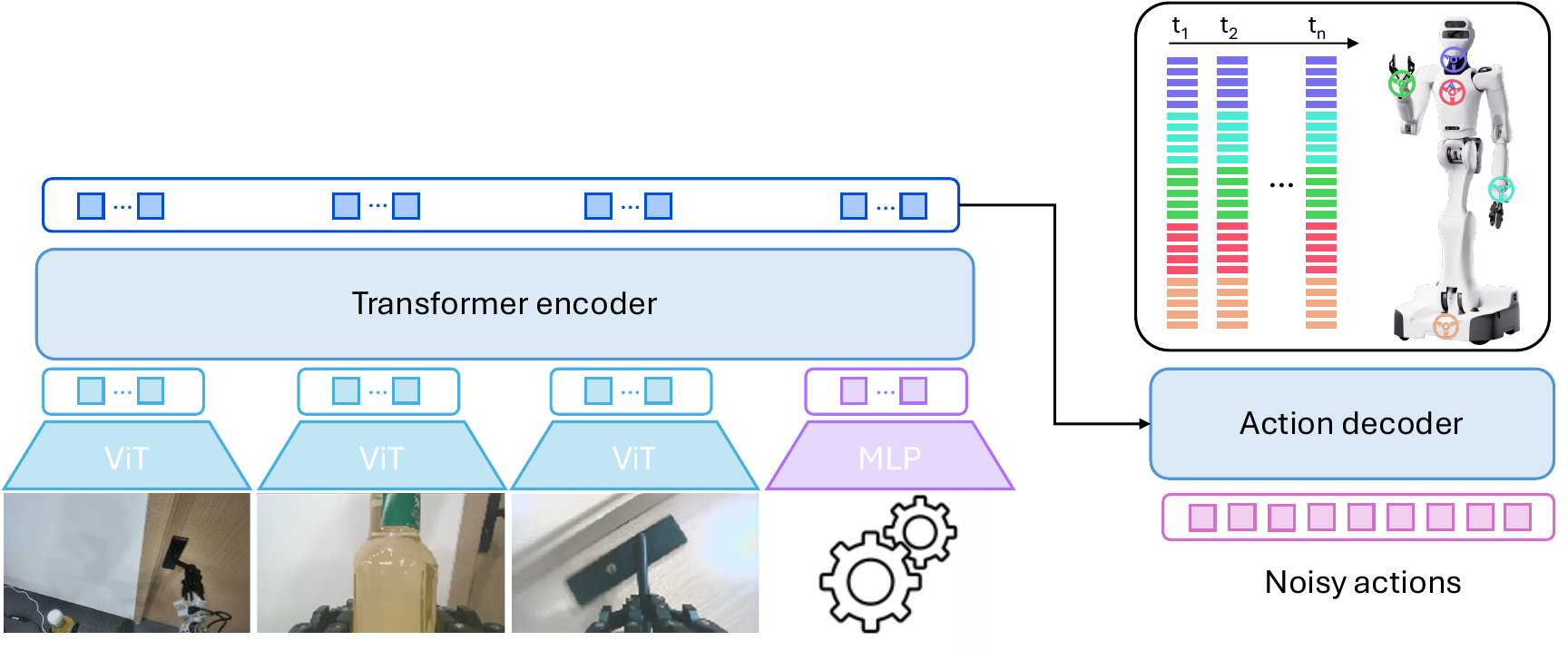}
    \caption{\textbf{\algoname model architecture for imitation learning.} \algoname denoises whole-body actions within the embodiment space and facilitates effective whole-body policy learning by directly predicting delta end-effector (EE) poses relative to the EE frame.}
    \label{fig:model_arch}
\end{figure*}

This section presents a simple yet effective approach for learning whole-body visuomotor policies. We begin by outlining preliminaries, then introduce \algoname, an algorithm designed to learn whole-body control policies from demonstration data. The method emphasizes simplicity and scalability, ensuring compatibility with large-scale pretraining. \algoname is trained using teleoperated demonstration data collected through the \fullname teleoperation interface. An overview of the model architecture is shown in Fig.~\ref{fig:model_arch}.

\subsection{Preliminaries}
\paragraph{Denoising Diffusion for Policy Learning}
Diffusion models~\cite{ddpm,improved_ddpm} learn to estimate the denoising score $\nabla_\mathbf{x}\log p_{\text{data}} (\mathbf{x})$ by adding noise to clean data $\mathbf{x} \sim p(\mathbf{x})$ (forward process) and learning to reverse the added noise (backward process). Noising the data distribution to isotropic Gaussian is performed in $T$ timesteps, with a pre-defined noising schedule $\alpha_t \in (0,1)$ and $\bar{\alpha}_t \coloneqq {\prod^t_{s=1}\alpha_s}$, according to:
\begin{align*}
\mathbf{z}_{t} =\sqrt{\bar{\alpha}_t} \mathbf{x} + \sqrt{1-\bar{\alpha}_{t}}\boldsymbol{\epsilon}, \text{ where } \boldsymbol{\epsilon} \sim \mathcal{N}(\mathbf{0}, \mathbf{I}).
\end{align*}
In the training process, the diffusion models $\epsilon_\phi (\mathbf{z}_t\vert t)$ learn to estimate the noise by
\begin{align*}
\mathcal{L}_{t} = \mathbb{E}_{\mathbf{x}, \boldsymbol{\epsilon} \sim \mathcal{N}(\mathbf{0},\mathbf{I}) }\left[\left\|\boldsymbol{\epsilon}_{\phi}\left(\mathbf{z}_t \vert t\right) - \boldsymbol{\epsilon} \right\|^{2}_{2}\right].
\end{align*}
Once trained, one can estimate $\mathbf{x}$ from noisy input and the corresponding noise prediction. 
Recently, DDPMs have been widely utilized to model policies $\pi_\phi$, conditioned on robot observations $\mathbf{O}_t$ , where the denoising network $\epsilon_\phi ( \mathbf{z}_t \vert \mathbf{O}_t, t)$ is trained through behavior cloning~\citep{chi2023diffusion,ze20243d,black2024pi_0,jiang2025behavior}.

\subsection{Whole-Body Visuomotor Policy}
We propose \algoname, a transformer-based model~\citep{vaswani2017attention} designed to learn coordinated whole-body actions for general manipulation tasks. As illustrated in Fig.~\ref{fig:model_arch}, 
\algoname performs denoising of whole-body actions in the space  end-effectoofr (EE) poses, rather than in the high-dimensional joint configuration space,  thereby mitigating error accumulation. Crucially, the model learns the delta poses in the egocentric frame of each end-effector, which promotes alignment between visual observations and action targets, and enhances robustness to large viewpoint variations commonly encountered in whole-body activities. The policy is trained using only RGB observations, ensuring seamless compatibility with large-scale Vision-Language-Action (VLA)~\citep{driess2023palm, zitkovich2023rt,kim2024openvla,black2024pi_0,liu2024rdt,team2025gemini, bjorck2025gr00t,intelligence2025pi_} model pretraining pipelines.

\paragraph{Model Architecture}
We sample image-state pairs and the corresponding action sequences from teleoperation demonstrations, with the action sequences as prediction targets. Observation images are captured from the head, left-hand, and right-hand cameras, with an input resolution of 224×224×3. These images are encoded using a ResNet-Small backbone consisting of four stages, each comprising two residual blocks. The final feature map has a resolution of 7×7×512. A 1×1 convolution is then applied to reduce the channel dimension to 384, followed by flattening to obtain 49 patch tokens per image. To encode camera-specific spatial information, we add a set of learnable positional embeddings to the tokens from each image. These embeddings are shared across all tokens within the same image to indicate both spatial position and camera source. In parallel, the robot's state - including base motion, torso configuration, end-effector poses of both arms, and head pose - is encoded into a state vector, which is projected via a linear layer into a 384-dimensional state token. A dedicated learnable positional embedding is added to denote the token's modality. Patch tokens from the three camera views (3×49 tokens) are concatenated with the state token to form a sequence of 148 tokens, which are fed into a Transformer encoder to extract condition features.
We then apply a conditional diffusion model to denoise the action sequence. The noisy action sequence is first mapped into the embedding space through a fully connected layer, forming a sequence of query tokens, each with a fixed timestep positional embedding. The decoder is a Transformer architecture with cross-attention, where the query tokens attend to the encoded condition tokens. The decoder output is then projected back to the original action space via a linear layer, yielding a denoised action sequence supervised by the training target.
The action sequence includes base control commands, torso movements, end-effector trajectories for both arms, and head control signals.

\paragraph{Real-time Trajectory Generation Module (RTG) }
Contemporary policy models typically generate a sequence of future actions conditioned on the current observation, where each sequence is referred to as an action chunk. While these action chunks are associated with temporal indices, their dynamic execution often introduces several challenges: (1) Intra-chunk jitter: inconsistency between consecutive actions within the same chunk; and (2) Inter-chunk discontinuities: 
abrupt transitions between different action chunks result in trajectory discontinuity which become more pronounced for whole-body manipulation, where the observations can vary significantly over time. Abrupt transitions in the action stream not only degrade execution stability, but may also pose risks to hardware integrity.

Our proposed approach is a lightweight post-processing module that can be seamlessly integrated with any visuomotor learning algorithm, provided the following conditions are satisfied: (1) the policy model generates one action chunk per inference cycle, with each action in the chunk associated with a timestamp aligned to the temporal resolution of the training data; and (2) the inference time is shorter than the execution duration of a single action chunk - ideally no more than half - to ensure sufficient overlap for smooth trajectory transitions. 

When a new action chunk is generated, observation timestamp associated with the new chunk is set to $t=0$. We denote the total duration of the chunk as $t_e$, the inference latency as $t_1$, and the processing time of the trajectory generation module as $t_2$. The newly generated action chunk is represented as $A_{new}[0:t_e]$, while the trajectory currently being executed is denoted by $A_{old}$. By the time the new action chunk becomes available, a duration of $t_1 + t_2$ has already elapsed since the observation timestamp $t=0$. The system then proceeds as follows: (1) the outdated portion $A_{new}[0: t_1 + t_2]$ is discarded; (2) execution continues with $A_{old}[0, t_1+t_2]$; and (3) starting from $t=t_1 + t_2$, a smooth blending is performed between $A_{new}[t_1+t_2: t_e]$ and $A_{old}[t_1+t_2: t_f]$, where $t_f$ is a predefined blending horizon (typically determined by the chunk duration and inference delay), with $t_f \leq$ the last timestamp of $A_{old}$.

The blending process leverages quadratic programming (QP) optimization to generate a smooth and dynamically feasible trajectory, comprehensively addressing continuity, transition quality, and physical constraints. The key components of this formulation include:
\begin{enumerate}
    \item \textbf{Smoothing term ($S_1$):} Minimize trajectory accelerations to ensure smooth motion, promoting natural transitions and reducing mechanical stress.
    \item \textbf{Deviation from the old trajectory ($S_2$):} Penalize deviation from the currently executing trajectory $A_{old}$ using a time-decaying weight $W_1(t)$, which decreases exponentially to reduce its influence as blending progresses.
    \item \textbf{Deviation from the new action chunk ($S_3$):} Encourage alignment with the newly generated action chunk $A_{new}$ using weight $W_2(t)=1-W_1(t)$, which increases over time to smoothly shift control authority to the new policy output.
    \item \textbf{Velocity constraint ($C_1$):} Enforce joint velocity limits to ensure the generated trajectory adheres to hardware safety and performance specifications.
\end{enumerate}
This QP-based approach enables real-time trajectory refinement, ensuring that transitions between action chunks are accurate, stable, and safe. For the initial action chunk - where no prior trajectory is available for blending - the trajectory is generated by optimizing over the chunk itself using the following components: smoothing term ($S_1$), deviation loss ($S_3$) with fixed weight $W_1(t)=1$, and velocity constraint $C_1$. Finally, the optimized discrete trajectory is interpolated using splines to ensure smooth motion and executed asynchronously with respect to inference, enabling continuous and stable control.

RTG offers key advantages in compatibility, comprehensiveness, and robustness. As a model-agnostic post-processing approach, it can be seamlessly integrated with any trained policy. It effectively mitigates both inter-chunk discontinuities and intra-chunk jitter, while explicitly accounting for hardware constraints such as joint velocity limits. Requiring only timestamped action chunks as input, RTG does not depend on precise estimation of policy inference time or frequency, making it inherently resilient to communication latency and inference delays.

\paragraph{Training and Deployment Details}
\algoname is trained to predict the added noise from noisy whole-body actions with the loss $\mathcal{L} = MSE(\boldsymbol{\epsilon}^t, \boldsymbol{\epsilon}_\theta(\cdot \vert t))$, where $\boldsymbol{\epsilon}^t$ represents the ground-truth added noise and $\boldsymbol{\epsilon}_\theta$ is the predicted noise.
During deployment, inference is performed on a workstation equipped with an NVIDIA RTX 4090 GPU, achieving an effective latency of \SI{0.05}{\second}.
Policy inference runs at \SI{20}{\hertz}, with the inferred results synchronously fed into the RTG at the same frequency.
The RTG then issues action commands to the controller at \SI{250}{\hertz}.
Therefore, a new policy action sequence is generated and the RTG state updated every \SI{0.05}{\second}, while the controller’s commands are refreshed every \SI{0.004}{\second}.

\begin{table}[]
\centering
\begin{tabular}{|c|c|c|l|c|}
\hline
\multicolumn{1}{|c|}{Task Name}   & \# demos             & \begin{tabular}[c]{@{}c@{}}Overall\\ SR \end{tabular} & \multicolumn{1}{c|}{Subtasks}                                                                & \begin{tabular}[c]{@{}c@{}}Subtask\\ SR\end{tabular} \\ \hline
\multirow{2}{*}{deliver a drink}  & \multirow{2}{*}{110} & \multirow{2}{*}{13/15}                                         &   open the door  \faWalking                                                                          & 14/15                                                          \\ \cline{4-5} 
                                  &                      &                                                                & \begin{tabular}[c]{@{}l@{}}enter the room \faWalking  \\ and place the drink on the table\end{tabular}   & 13/14                                                          \\ \hline
\multirow{2}{*}{store cat food}   & \multirow{2}{*}{50}  & \multirow{2}{*}{19/20}                                         & \begin{tabular}[c]{@{}l@{}}pick up the cat food\\ and place it into the cabinet\end{tabular} & 19/20                                                          \\ \cline{4-5} 
                                  &                      &                                                                & close the cabinet door                                                                       & 19/19                                                          \\ \hline
\multirow{3}{*}{throw away trash} & \multirow{3}{*}{100} & \multirow{3}{*}{13/30}                                         & press open the trash bin lid \faWalking                                                                  & 15/30                                                          \\ \cline{4-5} 
                                  &                      &                                                                & throw the paper cup into the trash bin                                                       & 15/15                                                          \\ \cline{4-5} 
                                  &                      &                                                                & close the trash bin lid                                                                      & 13/15                                                          \\ \hline
\multirow{2}{*}{organize shoes}   & \multirow{2}{*}{200} & \multirow{2}{*}{16/20}                                         & pick up shoes with both hands \faWalking                                                                 & 17/20                                                          \\ \cline{4-5} 
                                  &                      &                                                                & place the shoes on the shoe rack \faWalking                                                              & 16/17                                                          \\ \hline
\multirow{2}{*}{throw a toy}      & \multirow{2}{*}{74}  & \multirow{2}{*}{20/20}                                         & pick up the toy off the ground                                                               & 20/20                                                          \\ \cline{4-5} 
                                  &                      &                                                                & throw the toy far away (\textgreater{}2m)                                                            & 20/20                                                          \\ \hline
\multirow{3}{*}{pick up toys}     & \multirow{3}{*}{200} & 19/20                                                          & pick up the toy with right hand                                                              & 19/20                                                          \\ \cline{3-5} 
                                  &                      & \multirow{2}{*}{16/20}                                         & pick up the toy with left hand                                                               & 19/20                                                          \\ \cline{4-5} 
                                  &                      &                                                                & \begin{tabular}[c]{@{}l@{}}deliver the toy\\ from the left to the right hand\end{tabular}    & 16/19                                                          \\ \hline
\end{tabular}
\caption{\textbf{Task Success Rates.} For each task, we perform 15–30 evaluations, terminating the episode upon the first subtask failure. We report both per-subtask success rates and overall end-to-end task success rates. ``SR'' stands for success rates. Subtasks marked with the ~\faWalking~ icon indicate mobile manipulation.}
\label{tab:main_results}
\end{table}
\section{Experiments}

\subsection{Experiment Settings}
We evaluate \fullname on six representative tasks, as illustrated in Fig.~\ref{fig:teaser}. \textbf{(1) Deliver a drink} evaluates the system's ability to execute long-horizon tasks involving mobile manipulation and dexterous interaction with articulated objects, such as door handles. \textbf{(2) Store cat food} tests fine manipulation within confined spaces (e.g., low cabinets), coordinated bimanual actions under spatial constraints, and dynamic stability when handling heavy payloads (approximately 2kg). \textbf{(3) Throw away trash} challenges the robot with multi-stage bimanual coordination and precise manipulation of articulated components, such as lifting a trash bin lid. \textbf{(4) Organize shoes} emphasizes low-height whole-body coordination and synchronized bimanual manipulation. \textbf{(5) Throw a toy} highlights the robot's ability to perform dynamic whole-body movements and accurately grasp small objects from the floor. Finally, \textbf{(6) Pick up toys} serves as a comprehensive benchmark for sustained, accurate multi-object manipulation through coordinated bimanual control. These long-horizon, multi-stage tasks can be decomposed into a series of subtasks. For each task, we conduct 15–30 evaluation trials, terminating an episode upon the first subtask failure. We report both per-subtask success rates and overall end-to-end task success.

\begin{figure*}[t]
    \centering
    \includegraphics[width=\textwidth]{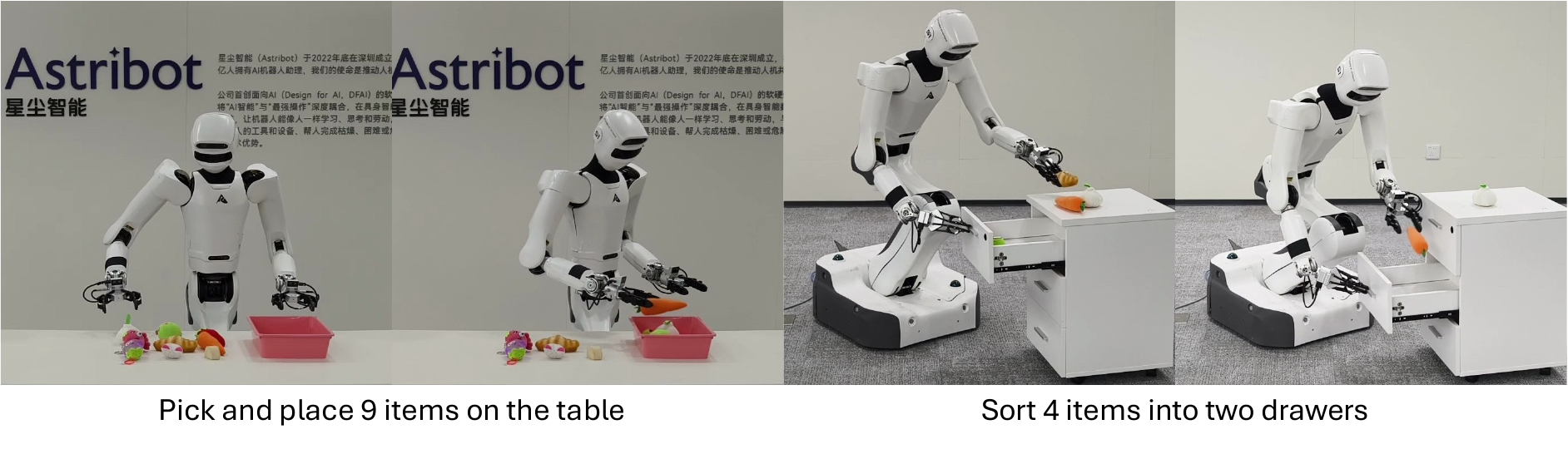}
    \caption{\textbf{Teleoperation tasks.} We evaluate teleoperation efficiency on two representative tasks. The first involves organizing nine items on a tabletop, demanding extensive reachability. The second requires organizing four items into two drawers at different heights, necessitating precise manipulation for opening and closing floor-level drawers.} 
    \label{fig:teleoperation}
\end{figure*}

\begin{table}[]
\centering
\begin{tabular}{|c|c|c|c|}
\hline
Task                                         & Human & \begin{tabular}[c]{@{}c@{}}Expert\\ Teleoperation\end{tabular} & \begin{tabular}[c]{@{}c@{}}Non-Expert\\ Teleoperation\end{tabular} \\ \hline
Pick and place 9 items on the table & 8.42                                                         & 10.8             & 15.93                                                \\ \hline
Sort 4 items into two drawers   & 7.12                                                            & 10.07          & 16.82                                                \\ \hline
\end{tabular}
\caption{\textbf{Comparison of task completion time.} Comparison between the time required by a human and the time taken by both an expert and a non-expert using our teleoperation interface to complete two representative tasks. All completion times are measured in seconds.}
\label{tab:teleopeartion_efficiency}
\end{table}

\subsection{\fullname Teleoperation Interface is Intuitive and Efficient for Policy Learning }\label{sec:exp:teleoperation}
We evaluate the teleoperation system of \fullname in terms of efficiency, and the quality of collected demonstrates for downstream policy learning. 
Specifically, we compare the time required for a human to complete a task with the time taken by both an expert and a non-expert using the \fullname teleoperation interface, across two representative tasks as demonstrated in Fig.~\ref{fig:teleoperation}. 
As shown in Tab.~\ref{tab:teleopeartion_efficiency}, for a simple tabletop task, the expert incurs an average $28.27\%$ overhead compared to direct human operation, while the non-expert incurs an additional $60.93\%$ overhead. For a more complex whole-body task - opening two drawers, sorting four items into them, and then closing the drawers - the expert's overhead increases to $41.43\%$, while non-expert incurs an additional $94.80\%$ overhead. These results demonstrate that the teleoperation interface enables efficient and intuitive operation, even in challenging whole-body manipulation scenarios. The intuitive teleoperation interface enables users to generate smooth and highly accurate robot actions. 
The tracking errors are remarkably low, as exemplified in the task of “pick up toys” from the ground using the right arm (Fig.~\ref{fig:tracking_error}), where data for six randomly selected joints are shown. The low tracking errors highlight the high fidelity of the collected demonstrations. Moreover, the teleoperation interface consistently provides high-quality data, with a $\sim 100\%$ success replay rates across diverse tasks.

\begin{figure*}[t]
    \centering
    \includegraphics[width=\textwidth]{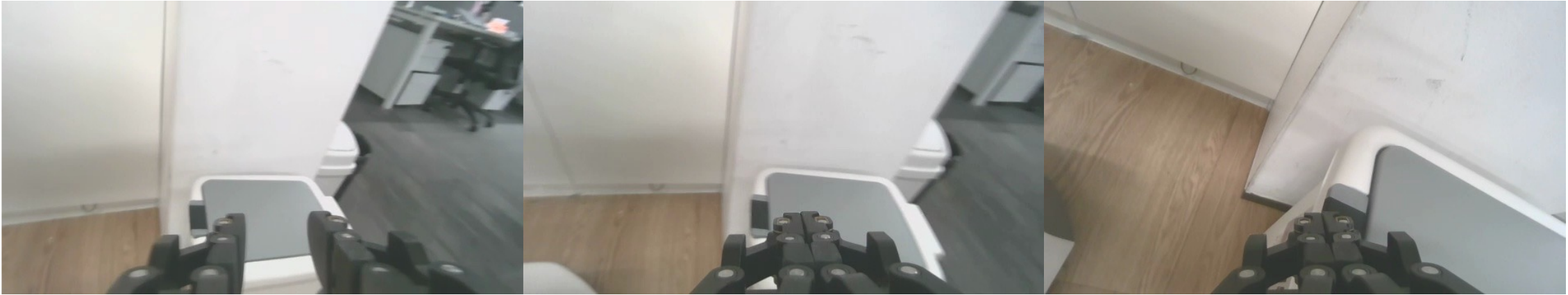}
    \caption{\textbf{Failure analysis.} The right-hand camera view reveals that precise pressing requires the gripper to be accurately aligned with the button's edge, which lacks distinctive visual features. } 
    \label{fig:press_failure}
\end{figure*}

\subsection{\algoname Is a Simple yet Effective Learning Method for Whole-body Activities}
We report the success rates of six representative tasks in  Tab.~\ref{tab:main_results}, including both per-subtask success rates and overall end-to-end task success rates.  Subtasks marked with the ~\faWalking~ icon indicate mobile manipulation, where the policy must successfully infer both navigation and manipulation commands.
As shown in Tab.~\ref{tab:main_results}, \algoname achieves an average success rate of 80\% and a  peak success rate of 100\%. The subtask that exhibits the lowest success rate is ``press open the trash bin lid''. Analysis of the action sequence and visual input in Fig.~\ref{fig:press_failure} reveals that precise pressing requires the gripper to align well with the edge of the button. However, the button appears small in the right-hand camera view, and shares a similar color with the gripper, resulting in weak visual contrast and limited perceptual cues. This task also demonstrates a stronger reliance on multi-view observations and is unable to fully exploit the benefits of the egocentric frame under the right-hand camera, ultimately hindering precise control.

\paragraph{Mitigate Error Accumulation by Learning in the End-Effector Space}
We compare two designs that differ only in their action representations: joint space versus end-effector (EE) space. We represent end-effector orientation using SO(3), as it provides a continuous representation which facilitates more efficient learning for neural networks~\citep{geist2024learning}.
For tasks with limited mobility where the robot torso remains fixed, policies trained with both representations perform comparably. However, in whole-body tasks, joint-based policies exhibit substantially lower precision compared to EE-based policies. This performance gap arises from the nature of whole-body control: achieving a desired end-effector pose requires coordinated motion across a hierarchical kinematic chain, spanning the mobile base, torso, and arms. Under the joint representation, the model must predict the positions of all joints in this chain; inaccuracies in proximal joints - particularly near the base - tend to propagate through the system, ultimately compromising end-effector accuracy.

This accumulation of error becomes particularly problematic in tasks that involve long error propagation paths, such as highly dynamic whole-body activities. To investigate this, we evaluate two representative tasks: a tabletop object-clearing task requiring only dual-arm coordination, and a ground-level object-sorting task demanding full-body coordination. Using 100 training demonstrations, the joint-space policy achieves a success rate of 18/20 on the tabletop task, while the end-effector (EE) space policy performs similarly with a 19/20 success rate. However, in the more challenging whole-body sorting task, the joint-space policy succeeds in only 5/20 trials, whereas the EE-space policy significantly outperforms it with an 18/20 success rate. These results underscore the superior generalization and control accuracy afforded by EE-space representations, particularly for complex whole-body tasks. Moreover, the EE representation is inherently more robust to variations in joint dynamics or motion tracking accuracy, such as those arising from differences between training and deployment robots, thereby mitigating performance degradation in real-world deployment.

\paragraph{Learning with Delta Action Representation Improves Trajectory Smoothness }
We observe that policies trained with absolute action representation often exhibit noticeable discontinuities between action steps during inference. This issue becomes particularly pronounced when increasing execution speed - either by raising the control frequency or the action magnitude - manifesting as abrupt arm motions, pauses, or oscillations in the trajectory, significantly undermining task stability and safety. We compare policies trained with delta action representation against those trained with absolute action representation, focusing on action continuity across two tasks: a tabletop object clearing task and a ground-level object sorting task.

\begin{figure*}[t]
    \centering
    \includegraphics[width=\textwidth]{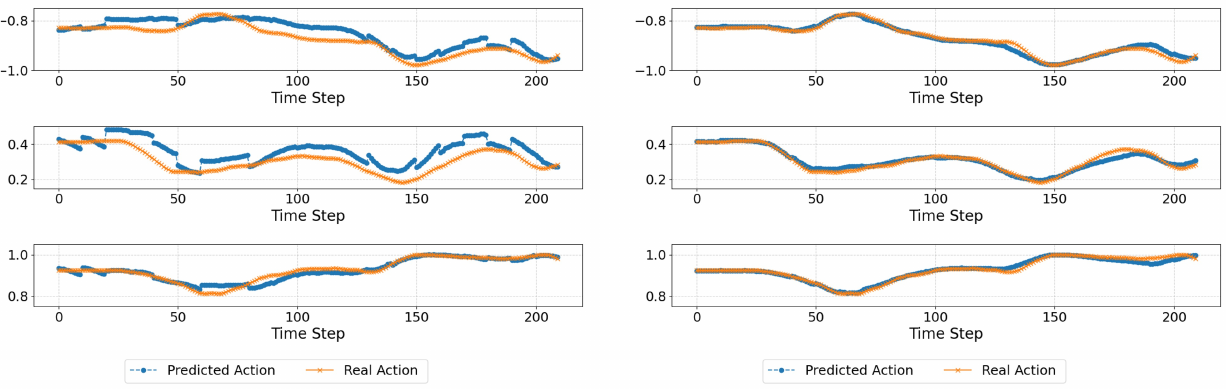}
    \caption{\textbf{Visualization of policy trajectories trained with different action representations.} Shown are action values of six randomly selected relevant joints. The policy trained with delta action representation (Right) produces trajectories that are noticeably smoother and more consistent compared to those generated by the absolute-action policy (Left).}
    \label{fig:absolute_vs_delta}
\end{figure*}

We evaluate all policies on a held-out test set and analyze relevant joint dimensions. As shown in Fig.~\ref{fig:absolute_vs_delta}, trajectories generated by the delta-action policy exhibit substantially smaller discontinuities between predicted action chunks across all dimensions, compared to those from the absolute-action policy. Quantitatively, the average per-step change across the entire trajectory is 0.0058 for the absolute-action policy and 0.0034 for the delta-action policy. Focusing specifically on transitions between action chunks, the absolute-action policy exhibits an average change of 0.0196, whereas the delta-action policy maintains a significantly lower average of 0.0032. These results demonstrate the superior smoothness and temporal consistency achieved by training policies with the delta action representation.

\begin{table}[]
\centering
\begin{tabular}{|c|c|c|c|}
\hline
\multicolumn{1}{|c|}{Task Type}                                                 & \multicolumn{1}{c|}{Dominant View} & \multicolumn{1}{c|}{Egocentric Delta} & \multicolumn{1}{c|}{Robot Delta} \\ \hline
\begin{tabular}[c]{@{}c@{}}Fine Grasping\\(Pen Pick-up)\end{tabular}              & Wrist Camera                                & \textbf{19 / 20}                                      & 17 / 20                                        \\ \hline
\begin{tabular}[c]{@{}c@{}}Large-Range Motion\\(Move Box)\end{tabular}            & Head Camera                                 & 17 / 20                                      & \textbf{19 / 20}                                        \\ \hline
\begin{tabular}[c]{@{}c@{}}Whole-Body Coordination\\(Ground Pick-up)\end{tabular} & Multi-View Fusion                           & \textbf{19 / 20}                                      & 16 / 20                                        \\ \hline
\end{tabular}
\caption{Policies trained with egocentric delta action representation achieves higher average success rate across different tasks.}
\label{tab:ee_frame_comparison}
\end{table}
\paragraph{Learning with Egocentric Delta Action Improves Observation-Action Alignment}
We represent delta actions as pose transformation matrices and investigate two distinct reference frames: the robot frame and the end-effector's egocentric frame, where deltas are defined relative to the end-effector's local coordinate system. We systematically compare these two representations across a variety of tabletop tasks, including fine manipulation tasks dominated by the wrist camera, large-range end-effector repositioning guided by the head camera, and ground-level object retrieval tasks that require full-body coordination. All evaluations are conducted under limited-data conditions using only 40 demonstrations per task. As summarized in Tab.~\ref{tab:ee_frame_comparison}, policies trained with egocentric delta action representation achieves higher success rates on average:
\begin{itemize}
    \item In tasks dominated by wrist camera observations, defining the delta in the end-effector's own frame is equivalent to modeling action targets directly in the camera frame. This tight spatial coupling enhances the alignment between visual observations and actions, which accelerates model convergence and improves motion accuracy.
    \item In tasks guided primarily by the head camera, the relative pose between the head camera and the robot's base frame is more stable. As a result, defining delta in the base frame facilitates a more consistent mapping between image features and control outputs, leading to better performance.
    \item For whole-body coordination tasks, the head camera viewpoint changes significantly during execution, which degrades the consistency between visual observations and the base frame. In contrast, defining delta in the end-effector frame offers a more stable and invariant representation of the motion target, resulting in better performance in complex, coordinated tasks.
    \item Additionally, we observe that under distributional shift - such as novel object positions trained with end-effector frame delta action representation exhibits stronger generalization in spatial reasoning and better deployment robustness, particularly in tasks requiring precise manipulation.
\end{itemize}

\begin{figure*}[t]
    \centering
    \includegraphics[width=\textwidth]{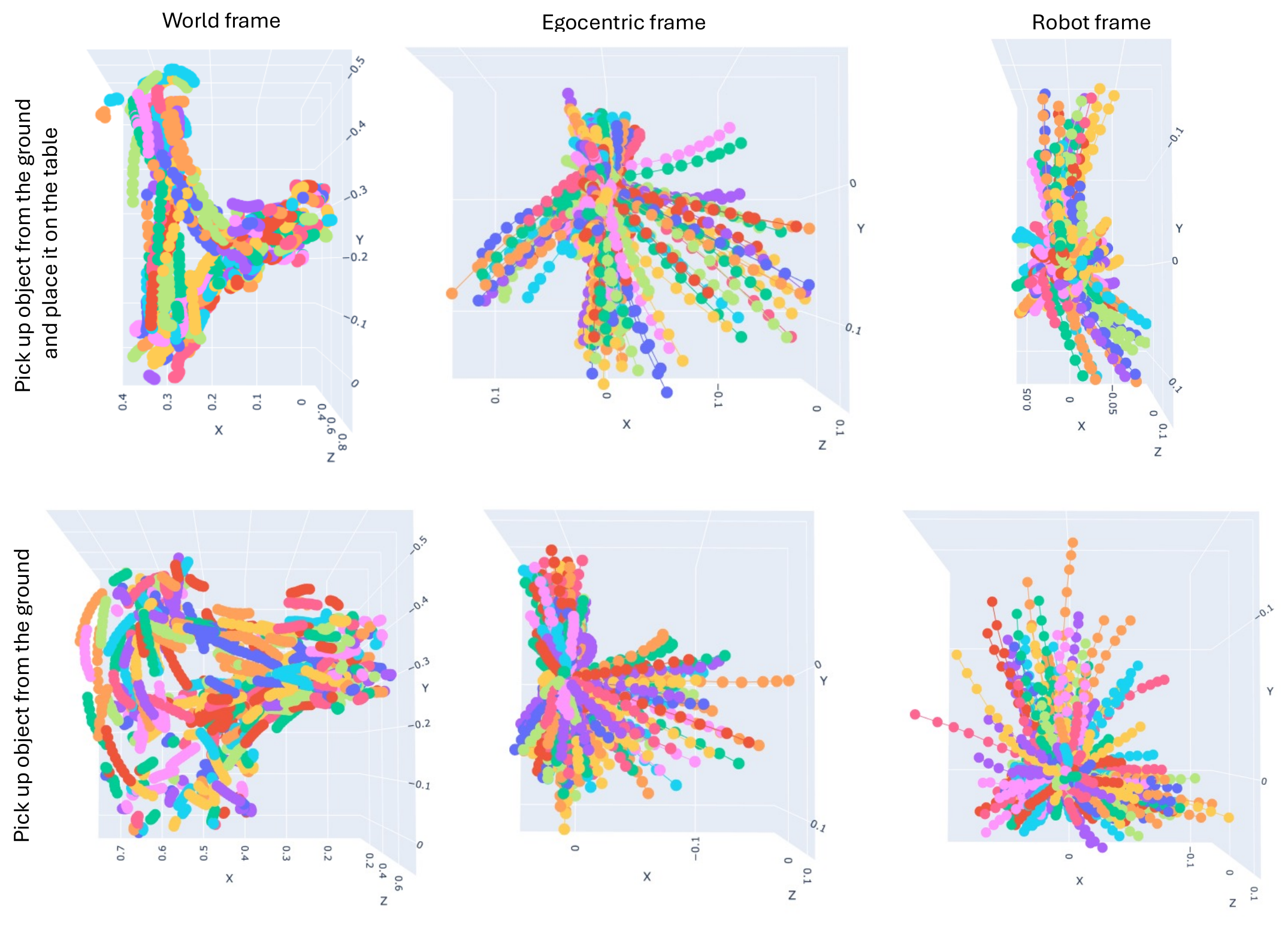}
    \caption{\textbf{Visualization of 500 trajectories under different action representations.} Egocentric-frame relative trajectories dynamically update the reference frame at each time step based on the current state of the end-effector. This leads to more structurally aligned and compact trajectory distributions across tasks. } 
    \label{fig:action_representations}
\end{figure*}

\begin{table}[]
\centering
\begin{tabular}{|c|c|c|c|c|}

\hline
Representation                                                               & \begin{tabular}[c]{@{}c@{}}Cross-Task\\ Consistency\end{tabular} & \begin{tabular}[c]{@{}c@{}}Trajectory \\ Complexity\end{tabular} & Adaptability & \begin{tabular}[c]{@{}c@{}}Control \\ Friendliness\end{tabular} \\ \hline
\begin{tabular}[c]{@{}c@{}}Absolute Trajectory\\  (World Frame)\end{tabular} & \faTimes~ (Poor)                                                           & \faTimes~ (High)                                                            &             \faTimes  &     \faTimes                                                            \\ \hline
\begin{tabular}[c]{@{}c@{}}Relative Trajectory\\ (Robot Frame)\end{tabular}  &  \faCheck~(Good)                                                           & \faCheck~(Moderate)                                               &    \faCheck           & \faCheck                                                                 \\ \hline
\begin{tabular}[c]{@{}c@{}}Relative Trajectory\\ (Egocentric)\end{tabular}   & \faLaugh~(Best)                                                           & \faLaugh~(Lowest)                                                         &  \faLaugh            &           \faLaugh                                                      \\ \hline
\end{tabular}
\caption{\textbf{Comparison of different action representations.}}
\vspace{-0.2em}
\label{tab:action_representation}
\end{table}

\paragraph{Discussion on Absolute and Relative Trajectory Representations}
In Fig.~\ref{fig:action_representations}, we visualize the end-effector trajectories of the right arm for two representative tasks: ``Pick up an object from the ground and place it on the table'' and ``Pick up objects from the ground'', each with 500 demonstration trajectories. We compare three primary forms of trajectory representation aforementioned: absolute trajectories in the world frame, relative trajectories in the robot frame, and relative trajectories in the egocentric frame.

Experimental results demonstrate that relative trajectory representations outperform absolute ones. Absolute trajectories, defined in the world coordinate frame, are highly sensitive to variations in task initial states, goal locations, and environment layouts. Consequently, the same action across different tasks results in diverse trajectories, lacking structural consistency and impeding policy generalization. In contrast, relative trajectories define actions in local coordinate frames that are more tightly coupled with the robot's state or the task context, enhancing structural consistency across tasks. \textbf{Robot-frame relative trajectories}, defined with respect to the robot base or the initial end-effector pose, improve trajectory consistency by filtering out global variations, making the trajectories more comparable and learnable. However, since the reference frame is fixed throughout the task, it may still be affected by robot movements or task-phase transitions, resulting in trajectory deformations. \textbf{Egocentric-frame relative trajectories} dynamically update the reference frame at each time step based on the current state of the end-effector, resulting in actions expressed from a consistent ``observer's perspective''. This leads to more structurally aligned and compact trajectory distributions across tasks. This representation exhibits the lowest variance, highest information density, and aligns naturally with the robot's natural perception and control, enabling more effective local feedback strategies. In summary, the comparative effectiveness of the three representations can be outlined as in Tab.~\ref{tab:action_representation}. We conclude that policies trained with relative trajectory representations in the egocentric frame achieve the best performance in multi-task manipulation learning. This representation not only minimizes inter-task variations but also significantly reduces the complexity of trajectory modeling, making it a more robust and efficient trajectory representation scheme.

\begin{figure*}[t]
    \centering
    \includegraphics[width=0.9\textwidth]{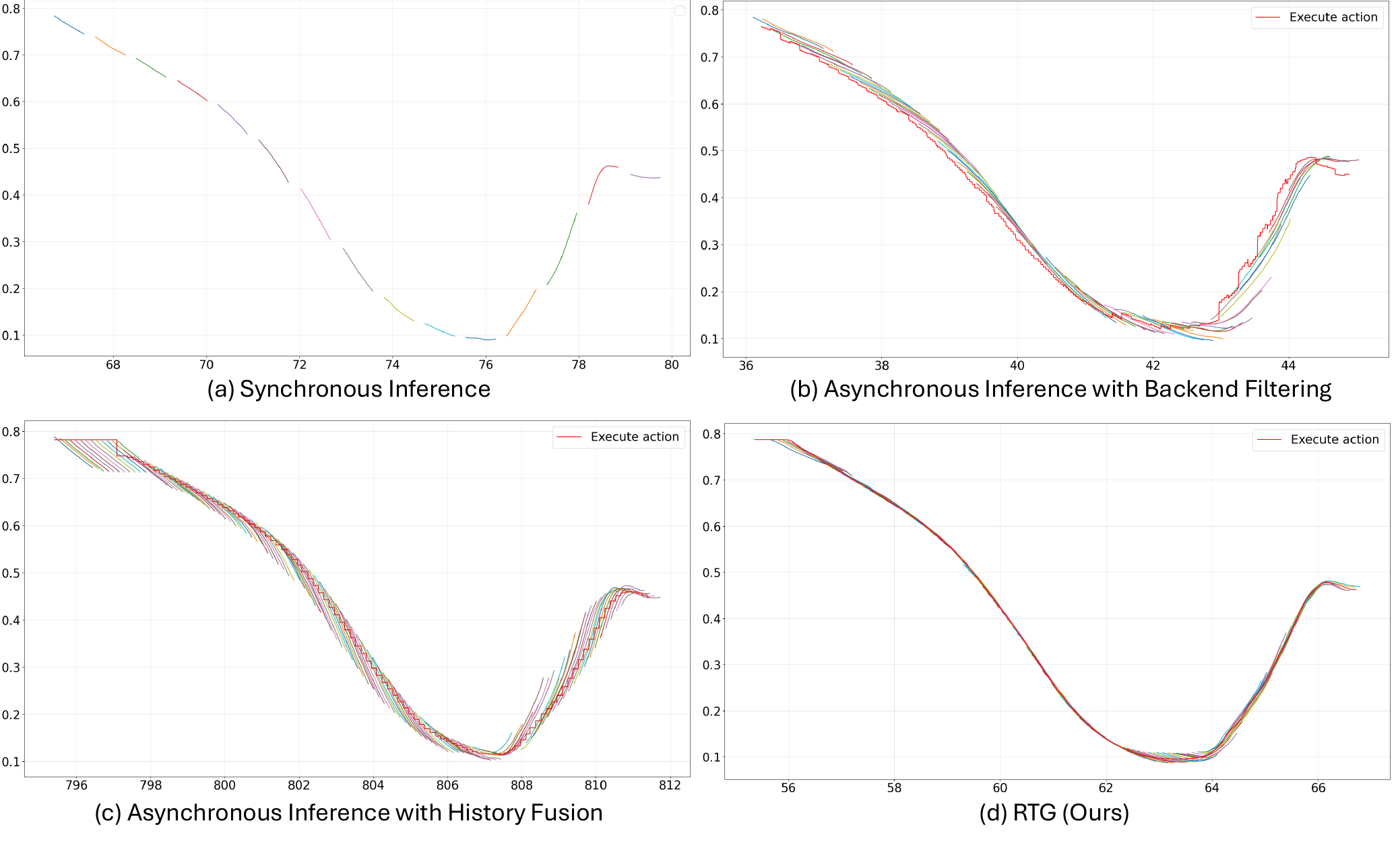}
    \caption{ \textbf{Comparison of action chuck smoothing methods on executed trajectories}. $x$ axis denotes time (s) and $y$ axis denotes action values over time. The multicolored lines represent action chunks generated by the VLA model, while the red line shows the robot's actual (pre-filtered) executed trajectory. Our method (RTG) effectively generates smooth trajectories that closely follow the predicted action chunks. } 
    \vspace{-0.5em}
    \label{fig:trajectory_generation}
\end{figure*}

\paragraph{Effectiveness of Real-time Trajectory (RTG) Generation Module}
We compare RTG with several commonly used strategies to mitigate discontinuities between action chunks. \textbf{Synchronous inference} executes the current action chunk (or a subset thereof) to completion before generating and executing the next, ensuring temporal consistency but introduces execution pauses due to inference latency, reducing responsiveness during transitions. \textbf{Asynchronous inference} immediately switches to a newly available action chunk, which usually requires filtering techniques to smooth out abrupt transitions. A more advanced technique, \textbf{Asynchronous inference with history fusion}, as used in ACT~\citep{Zhao2023LearningFB}, reduces discontinuities by fusing actions associated with the same timestamp across recent chunks through weighted averaging. This strategy incurs latency and can be unreliable when observations shift abruptly; reliance on outdated or inaccurate trajectories hampers timely transition to more suitable action chunks, ultimately degrading task success rates.

\begin{figure*}[t]
    \centering
    \includegraphics[width=0.9\textwidth]{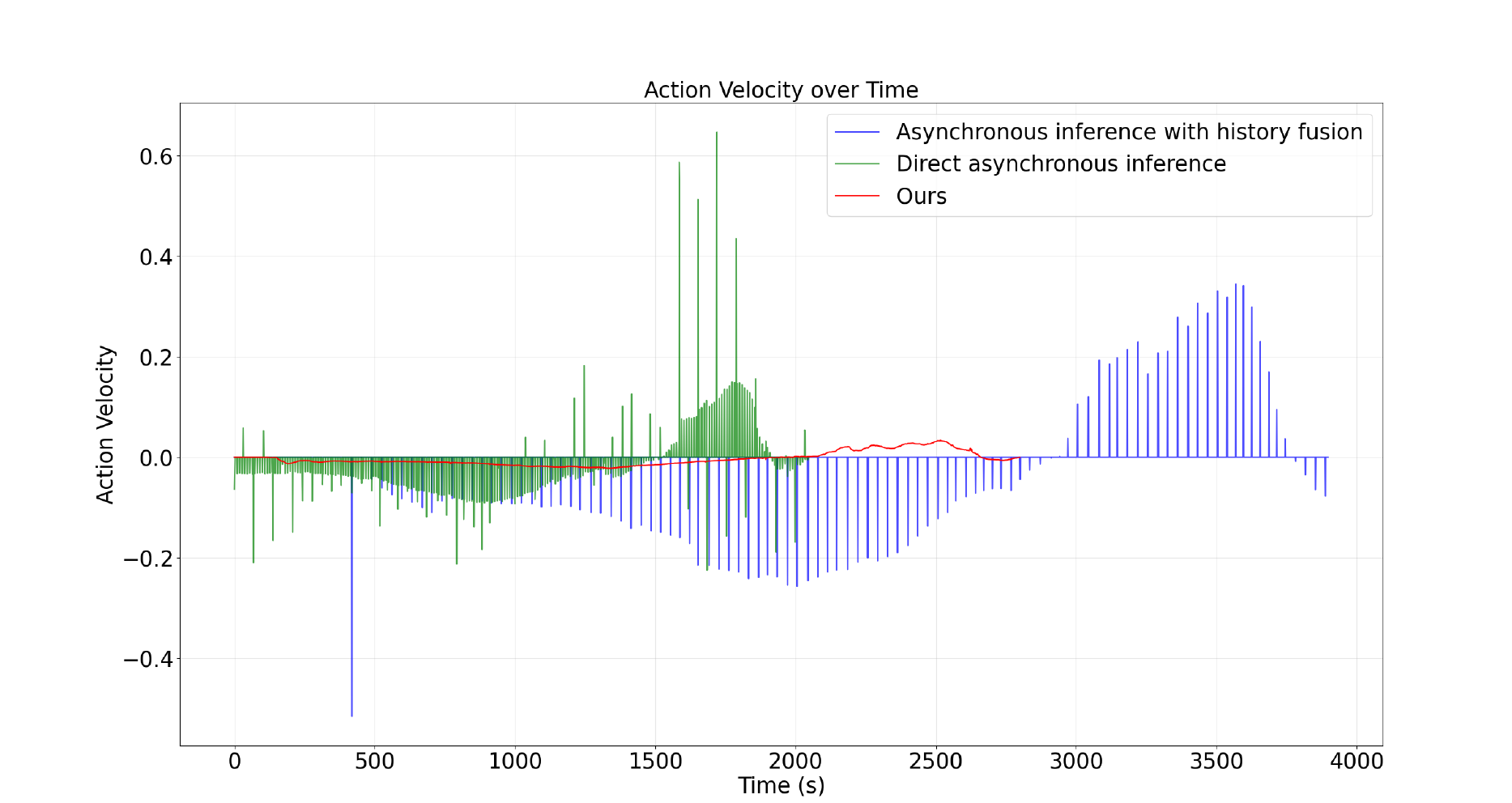}
    \caption{ \textbf{Comparison of action chuck smoothing methods on action velocities}. $x$ axis denotes time (s) and $y$ axis denotes action velocity over time. Our method (RTG) effectively constrains velocity within a safe range. } 
    \label{fig:trajectory_generation_velocity}
\end{figure*}

Evaluations are conducted using a whole-body VLA model with an action chunk size of 32, operating at an intended frequency of 10 Hz. After accounting for communication delays and blocking data waits, the effective inference frequency is reduced to approximately 7 Hz. For clarity, results are visualized using one-dimensional data. Fig.\ref{fig:trajectory_generation} illustrates trajectories from different methods: multicolored lines denote the action chunks predicted by the VLA model, while the red line represents the robot's actual (pre-filtered) executed trajectory. The corresponding action velocities are shown in Fig.~\ref{fig:trajectory_generation_velocity}. As shown in Fig.~\ref{fig:trajectory_generation} (a), synchronous inference introduces a noticeable pause at the end of each action chunk due to the blocking wait for the next inference cycle. Nevertheless, interpolation within the same chunk ensures smooth and continuous execution. Asynchronous inference incurs significant velocity spikes and abrupt transitions, without proper filtering, executing such discontinuous trajectories directly can jeopardize hardware safety. While asynchronous inference with history fusion alleviates some discontinuities, it still produces velocity magnitudes that require additional filtering. As shown in the velocity profile (Fig.\ref{fig:trajectory_generation_velocity}), our proposed method effectively constrains velocities within a safe operational bound (maximum 0.033 m/s), enabling smooth execution even in the absence of backend filtering. Moreover, the executed trajectory closely aligns with the predicted action chunks (Fig.~\ref{fig:trajectory_generation} (d)), reducing trajectory drift and minimizing out-of-distribution states - both critical for enhancing task success and policy robustness.

\section{Conclusion, Limitations, and Future Directions}
\label{sec:conclusion}

In this paper, we present \fullname{}, a robot learning suite for whole-body manipulation with three core components: (1) a safe robotic platform with human-level physical capabilities; (2) an intuitive and scalable teleoperation interface for real-world data collection; and (3) a simple yet effective whole-body visuomotor policy compatible with large-scale policy pretraining.

While demonstrating strong performance across a wide range of tasks, \fullname{} still faces certain limitations - particularly in learning to tackle more advanced tasks that demand greater agility, dexterity, or long-term memory.

In future work, we plan to strategically invest in the following directions:
\begin{itemize}
    \item 
    \textbf{Robotic hardware enhancement:} We will continue advancing our robotic platform to achieve higher capabilities and improved safety, aiming to support both real-world deployment and cutting-edge research, thereby pushing the boundaries of embodied intelligence.
    \item 
    \textbf{Human–robot interaction:} We will iterate on more intuitive and intelligent human–robot interaction methods, including but not limited to AI-assisted teleoperation and adaptive interfaces that facilitate seamless collaboration between humans and robots. 
    \item
    \textbf{Model and system scalability:} We will refine our model architectures and pretraining strategies to enhance both training and inference efficiency, with the long-term goal of enabling scalable learning systems and broad deployment in human-centered environments.
    
\end{itemize}

\section{Contributions}
\label{sec:contribution}
\begin{itemize}
    \item \textbf{Core Contributors:} Guang Gao, Jinbo Zuo
    \item  \textbf{Contributions:} Junnan Jiang, Jingfan Zhang, Xianwen Zeng, Yuejiang Zhu, Lianyang Ma, Ke Chen, Minhua Sheng, Ruirui Zhang
    \item  \textbf{Project Leads:} Jianan Wang, Zhaohui An
\end{itemize}


\clearpage
\begin{thebibliography}{27}
\providecommand{\natexlab}[1]{#1}
\providecommand{\url}[1]{\texttt{#1}}
\expandafter\ifx\csname urlstyle\endcsname\relax
  \providecommand{\doi}[1]{doi: #1}\else
  \providecommand{\doi}{doi: \begingroup \urlstyle{rm}\Url}\fi

\bibitem[Bjorck et~al.(2025)Bjorck, Casta{\~n}eda, Cherniadev, Da, Ding, Fan, Fang, Fox, Hu, Huang, et~al.]{bjorck2025gr00t}
J.~Bjorck, F.~Casta{\~n}eda, N.~Cherniadev, X.~Da, R.~Ding, L.~Fan, Y.~Fang, D.~Fox, F.~Hu, S.~Huang, et~al.
\newblock Gr00t n1: An open foundation model for generalist humanoid robots.
\newblock \emph{arXiv preprint arXiv:2503.14734}, 2025.

\bibitem[Black et~al.(2024)Black, Brown, Driess, Esmail, Equi, Finn, Fusai, Groom, Hausman, Ichter, et~al.]{black2024pi_0}
K.~Black, N.~Brown, D.~Driess, A.~Esmail, M.~Equi, C.~Finn, N.~Fusai, L.~Groom, K.~Hausman, B.~Ichter, et~al.
\newblock $\pi$0: A vision-language-action flow model for general robot control.
\newblock \emph{arXiv preprint arXiv:2410.24164}, 2024.

\bibitem[Chi et~al.(2023)Chi, Xu, Feng, Cousineau, Du, Burchfiel, Tedrake, and Song]{chi2023diffusion}
C.~Chi, Z.~Xu, S.~Feng, E.~Cousineau, Y.~Du, B.~Burchfiel, R.~Tedrake, and S.~Song.
\newblock Diffusion policy: Visuomotor policy learning via action diffusion.
\newblock \emph{The International Journal of Robotics Research}, page 02783649241273668, 2023.

\bibitem[Dai and Wang(2025)]{dai2025co}
Y.~Dai and J.~Wang.
\newblock Co-evolving embodied intelligence with design for artificial intelligence architecture.
\newblock \emph{Nature Reviews Electrical Engineering}, 2\penalty0 (3):\penalty0 149--150, 2025.

\bibitem[Dosovitskiy et~al.(2020)Dosovitskiy, Beyer, Kolesnikov, Weissenborn, Zhai, Unterthiner, Dehghani, Minderer, Heigold, Gelly, et~al.]{dosovitskiy2020image}
A.~Dosovitskiy, L.~Beyer, A.~Kolesnikov, D.~Weissenborn, X.~Zhai, T.~Unterthiner, M.~Dehghani, M.~Minderer, G.~Heigold, S.~Gelly, et~al.
\newblock An image is worth 16x16 words: Transformers for image recognition at scale.
\newblock \emph{arXiv preprint arXiv:2010.11929}, 2020.

\bibitem[Driess et~al.(2023)Driess, Xia, Sajjadi, Lynch, Chowdhery, Ichter, Wahid, Tompson, Vuong, Yu, Huang, Chebotar, Sermanet, Duckworth, Levine, Vanhoucke, Hausman, Toussaint, Greff, Zeng, Mordatch, and Florence]{driess2023palm}
D.~Driess, F.~Xia, M.~S.~M. Sajjadi, C.~Lynch, A.~Chowdhery, B.~Ichter, A.~Wahid, J.~Tompson, Q.~H. Vuong, T.~Yu, W.~Huang, Y.~Chebotar, P.~Sermanet, D.~Duckworth, S.~Levine, V.~Vanhoucke, K.~Hausman, M.~Toussaint, K.~Greff, A.~Zeng, I.~Mordatch, and P.~R. Florence.
\newblock Palm-e: An embodied multimodal language model.
\newblock In \emph{International Conference on Machine Learning}, 2023.

\bibitem[Geist et~al.(2024)Geist, Frey, Zhobro, Levina, and Martius]{geist2024learning}
A.~R. Geist, J.~Frey, M.~Zhobro, A.~Levina, and G.~Martius.
\newblock Learning with 3d rotations, a hitchhiker's guide to so (3).
\newblock \emph{arXiv preprint arXiv:2404.11735}, 2024.

\bibitem[Gupta et~al.(2021)Gupta, Savarese, Ganguli, and Fei-Fei]{gupta2021embodied}
A.~Gupta, S.~Savarese, S.~Ganguli, and L.~Fei-Fei.
\newblock Embodied intelligence via learning and evolution.
\newblock \emph{Nature communications}, 12\penalty0 (1):\penalty0 5721, 2021.

\bibitem[Haddadin et~al.(2017)Haddadin, De~Luca, and Albu-Sch{\"a}ffer]{haddadin2017robot}
S.~Haddadin, A.~De~Luca, and A.~Albu-Sch{\"a}ffer.
\newblock Robot collisions: A survey on detection, isolation, and identification.
\newblock \emph{IEEE Transactions on Robotics}, 33\penalty0 (6):\penalty0 1292--1312, 2017.

\bibitem[Ho et~al.(2020)Ho, Jain, and Abbeel]{ddpm}
J.~Ho, A.~Jain, and P.~Abbeel.
\newblock Denoising diffusion probabilistic models.
\newblock \emph{Advances in neural information processing systems}, 33:\penalty0 6840--6851, 2020.

\bibitem[Intelligence et~al.(2025)Intelligence, Black, Brown, Darpinian, Dhabalia, Driess, Esmail, Equi, Finn, Fusai, et~al.]{intelligence2025pi_}
P.~Intelligence, K.~Black, N.~Brown, J.~Darpinian, K.~Dhabalia, D.~Driess, A.~Esmail, M.~Equi, C.~Finn, N.~Fusai, et~al.
\newblock $\pi\_0.5$: a vision-language-action model with open-world generalization.
\newblock \emph{arXiv preprint arXiv:2504.16054}, 2025.

\bibitem[Jiang et~al.(2025)Jiang, Zhang, Wong, Wang, Ze, Yin, Gokmen, Song, Wu, and Fei-Fei]{jiang2025behavior}
Y.~Jiang, R.~Zhang, J.~Wong, C.~Wang, Y.~Ze, H.~Yin, C.~Gokmen, S.~Song, J.~Wu, and L.~Fei-Fei.
\newblock Behavior robot suite: Streamlining real-world whole-body manipulation for everyday household activities.
\newblock \emph{arXiv preprint arXiv:2503.05652}, 2025.

\bibitem[Jin et~al.(2024)Jin, Kobayashi, and Doi]{jin2024real}
T.~Jin, T.~Kobayashi, and M.~Doi.
\newblock Real-time detailed self-collision avoidance in whole-body model predictive control.
\newblock In \emph{2024 IEEE-RAS 23rd International Conference on Humanoid Robots (Humanoids)}, pages 675--681. IEEE, 2024.

\bibitem[Kim et~al.(2024)Kim, Pertsch, Karamcheti, Xiao, Balakrishna, Nair, Rafailov, Foster, Lam, Sanketi, et~al.]{kim2024openvla}
M.~J. Kim, K.~Pertsch, S.~Karamcheti, T.~Xiao, A.~Balakrishna, S.~Nair, R.~Rafailov, E.~Foster, G.~Lam, P.~Sanketi, et~al.
\newblock Openvla: An open-source vision-language-action model.
\newblock \emph{arXiv preprint arXiv:2406.09246}, 2024.

\bibitem[Liu et~al.(2023)Liu, Jiang, Zhao, and Mei]{liu2023collision}
B.~Liu, G.~Jiang, F.~Zhao, and X.~Mei.
\newblock Collision-free motion generation based on stochastic optimization and composite signed distance field networks of articulated robot.
\newblock \emph{IEEE Robotics and Automation Letters}, 8\penalty0 (11):\penalty0 7082--7089, 2023.

\bibitem[Liu et~al.(2024)Liu, Wu, Li, Tan, Chen, Wang, Xu, Su, and Zhu]{liu2024rdt}
S.~Liu, L.~Wu, B.~Li, H.~Tan, H.~Chen, Z.~Wang, K.~Xu, H.~Su, and J.~Zhu.
\newblock Rdt-1b: a diffusion foundation model for bimanual manipulation.
\newblock \emph{arXiv preprint arXiv:2410.07864}, 2024.

\bibitem[Nichol and Dhariwal(2021)]{improved_ddpm}
A.~Q. Nichol and P.~Dhariwal.
\newblock Improved denoising diffusion probabilistic models.
\newblock In \emph{International conference on machine learning}, pages 8162--8171. PMLR, 2021.

\bibitem[Oquab et~al.(2023)Oquab, Darcet, Moutakanni, Vo, Szafraniec, Khalidov, Fernandez, Haziza, Massa, El-Nouby, et~al.]{oquab2023dinov2}
M.~Oquab, T.~Darcet, T.~Moutakanni, H.~Vo, M.~Szafraniec, V.~Khalidov, P.~Fernandez, D.~Haziza, F.~Massa, A.~El-Nouby, et~al.
\newblock Dinov2: Learning robust visual features without supervision.
\newblock \emph{arXiv preprint arXiv:2304.07193}, 2023.

\bibitem[Qian et~al.(2018)Qian, Zi, Shang, and Xu]{qian2018review}
S.~Qian, B.~Zi, W.-W. Shang, and Q.-S. Xu.
\newblock A review on cable-driven parallel robots.
\newblock \emph{Chinese Journal of Mechanical Engineering}, 31\penalty0 (1):\penalty0 1--11, 2018.

\bibitem[Radford et~al.(2021)Radford, Kim, Hallacy, Ramesh, Goh, Agarwal, Sastry, Askell, Mishkin, Clark, et~al.]{radford2021learning}
A.~Radford, J.~W. Kim, C.~Hallacy, A.~Ramesh, G.~Goh, S.~Agarwal, G.~Sastry, A.~Askell, P.~Mishkin, J.~Clark, et~al.
\newblock Learning transferable visual models from natural language supervision.
\newblock In \emph{International conference on machine learning}, pages 8748--8763. PmLR, 2021.

\bibitem[Team et~al.(2021)Team, Abramson, Ahuja, Brussee, Carnevale, Cassin, Fischer, Georgiev, Goldin, Gupta, et~al.]{team2021creating}
D.~I.~A. Team, J.~Abramson, A.~Ahuja, A.~Brussee, F.~Carnevale, M.~Cassin, F.~Fischer, P.~Georgiev, A.~Goldin, M.~Gupta, et~al.
\newblock Creating multimodal interactive agents with imitation and self-supervised learning.
\newblock \emph{arXiv preprint arXiv:2112.03763}, 2021.

\bibitem[Team et~al.(2025)Team, Abeyruwan, Ainslie, Alayrac, Arenas, Armstrong, Balakrishna, Baruch, Bauza, Blokzijl, et~al.]{team2025gemini}
G.~R. Team, S.~Abeyruwan, J.~Ainslie, J.-B. Alayrac, M.~G. Arenas, T.~Armstrong, A.~Balakrishna, R.~Baruch, M.~Bauza, M.~Blokzijl, et~al.
\newblock Gemini robotics: Bringing ai into the physical world.
\newblock \emph{arXiv preprint arXiv:2503.20020}, 2025.

\bibitem[Vaswani et~al.(2017)Vaswani, Shazeer, Parmar, Uszkoreit, Jones, Gomez, Kaiser, and Polosukhin]{vaswani2017attention}
A.~Vaswani, N.~Shazeer, N.~Parmar, J.~Uszkoreit, L.~Jones, A.~N. Gomez, {\L}.~Kaiser, and I.~Polosukhin.
\newblock Attention is all you need.
\newblock \emph{Advances in neural information processing systems}, 30, 2017.

\bibitem[Ze et~al.(2024)Ze, Zhang, Zhang, Hu, Wang, and Xu]{ze20243d}
Y.~Ze, G.~Zhang, K.~Zhang, C.~Hu, M.~Wang, and H.~Xu.
\newblock 3d diffusion policy: Generalizable visuomotor policy learning via simple 3d representations.
\newblock \emph{arXiv preprint arXiv:2403.03954}, 2024.

\bibitem[Zhai et~al.(2023)Zhai, Mustafa, Kolesnikov, and Beyer]{zhai2023sigmoid}
X.~Zhai, B.~Mustafa, A.~Kolesnikov, and L.~Beyer.
\newblock Sigmoid loss for language image pre-training.
\newblock In \emph{Proceedings of the IEEE/CVF international conference on computer vision}, pages 11975--11986, 2023.

\bibitem[Zhao et~al.(2023)Zhao, Kumar, Levine, and Finn]{Zhao2023LearningFB}
T.~Zhao, V.~Kumar, S.~Levine, and C.~Finn.
\newblock Learning fine-grained bimanual manipulation with low-cost hardware.
\newblock \emph{ArXiv}, abs/2304.13705, 2023.
\newblock URL \url{https://api.semanticscholar.org/CorpusID:258331658}.

\bibitem[Zitkovich et~al.(2023)Zitkovich, Yu, Xu, Xu, Xiao, Xia, Wu, Wohlhart, Welker, Wahid, et~al.]{zitkovich2023rt}
B.~Zitkovich, T.~Yu, S.~Xu, P.~Xu, T.~Xiao, F.~Xia, J.~Wu, P.~Wohlhart, S.~Welker, A.~Wahid, et~al.
\newblock Rt-2: Vision-language-action models transfer web knowledge to robotic control.
\newblock In \emph{Conference on Robot Learning}, pages 2165--2183. PMLR, 2023.

\end{thebibliography}
\end{document}